\documentclass[letterpaper, 10 pt, conference]{ieeeconf}  

\IEEEoverridecommandlockouts                              

\usepackage{graphics} 
\usepackage{epsfig} 
\usepackage{mathptmx} 
\usepackage{times} 
\usepackage{amsmath, amssymb, bm} 
\usepackage{cmd_tex}
\usepackage{caption, subcaption}
\usepackage{xcolor}
\usepackage[T1]{fontenc}

\title{\LARGE \bf
Design of Multi-model Linear Inferential Sensors\\ with SVM-based Switching Logic}

\author{Martin Mojto, Miroslav Fikar and Radoslav Paulen
\thanks{*This research is funded by the Slovak Research and Development Agency under the project APVV-20-0261 and by the Scientific Grant Agency of the Slovak Republic under the grant VEGA 1/0691/21.}
\thanks{The authors are with Faculty of Chemical and Food Technology, Slovak University of Technology in Bratislava (STUBA), Radlinsk\'{e}ho 9, 912~37 Bratislava, Slovak Republic {\tt\small \{martin.mojto, miroslav.fikar, radoslav.paulen\}@stuba.sk}}%
}

\begin{document}
\maketitle
\thispagestyle{empty}
\pagestyle{empty}

\begin{abstract}
We study the problem of data-based design of multi-model linear inferential (soft) sensors. The multi-model linear inferential sensors promise increased prediction accuracy yet simplicity of the model structure and training. The standard approach to the multi-model inferential sensor design consists in three separate steps: 1)~data labeling (establishing training subsets for individual models), 2)~data classification (creating a switching logic for the models), and 3)~training of individual models. There are two main issues with this concept: a) as steps 2)\,\&\,3) are separate, discontinuities can occur when switching between the models; b) as steps 1)\,\&\,3) are separate, data labelling disregards the quality of the resulting model. Our contribution aims at both the mentioned problems, where, for the problem a), we introduce a novel SVM-based model training coupled with switching logic identification and, for the problem b), we propose a direct optimization of data labelling. We illustrate the proposed methodology and its benefits on an example from the chemical engineering domain.
\end{abstract}

\section{Introduction}
The use potential of inferential (soft) sensors is rising in many industrial and engineering fields. Frequent and accurate estimation of key variables plays a significant role in monitoring and control in various real-world use cases~\cite{abouzari_2021, li_2021, qi_2021}. One way of dealing with the estimation of such variables is represented by inferential (soft) sensors~\cite{joseph_1978}. The principle of inferential sensing is to infer the desired hard-to-measure variables according to other easy-to-measure variables (e.g., temperatures, pressures) in the process, e.g., see~\cite{qin_1997, zhu_2020}. The designed inferential sensor thus usually yields less expensive yet more frequent estimation of the key process variable than physical sensors.

The main design trade-off of the inferential sensors is that the higher accuracy is provided at the cost of the higher complexity of the model structure and its model training. The industrial processes are usually nonlinear, which often prohibits a use of simple (linear) inferential sensors due to their inaccuracy (i.e., poor extrapolation performance). This aspect can be easily compensated by the design of more complex inferential sensors, e.g., nonlinear inferential sensor~\cite{park_2000} or dynamic inferential sensor~\cite{wang_2019}. However, the design of more complex sensors usually involves much greater effort (model selection, data treatment, model validation, etc.) and there are even situations in practice when only linear inferential sensor can be implemented along with the present control solution. In such situations, it is possible to approximate the nonlinear behavior of process by designing a so-called multi-model inferential sensor (MIS)~\cite{khatibisepehr_2012}. The MISs found their use mostly in the same fields as standard inferential sensors, e.g., in the petrochemical industry~\cite{khatibisepehr_2012}, in manufacturing~\cite{zhongda_2016}, and in the process industry~\cite{hou_2020}.

The standard approach to the multi-model inferential sensor design consists in three separate steps: data labeling, data classification, and training of individual models.
Data labeling is used to establish training subsets for individual models and is usually done using some clustering method, e.g., $k$-means clustering~\cite{forgy_1965}. Data classification aims at creating a switching logic for the models. This step can be achieved using multi-class (linear) separation. A suitable method here is support vector machines (SVM)~\cite{vapnik_1992}, which splits the whole space into the desired number of model validity regions (classes). Linear hyperplanes are used for the switching logic due to their simplicity, interpretability, and ease of implementation. The final training of the individual models can be done within the particular regions using standard maximum-likelihood methods or using some dimensionality-reduction methods~\cite{mojto_2021}. In fact, the design of MIS is closely similar to the piece-wise affine identification~\cite{bako_2019}.

As the classification and training are performed separately, discontinuities can occur when switching between the models. Also the performance of the trained MIS depends strongly on the quality of the switching logic between the models, which stems from good data labeling. Simply speaking, data labelling disregards the quality of the resulting model. This contribution aims to overcome these drawbacks of the standard approach. We develop a new approach that effectively combines the training of the individual models with SVM. This approach ensures continuity of the designed models. Subsequently, we extend the MIS design method with direct optimal data labeling, thus avoiding the need for pre-classification of the data. We analyze the effectiveness of the aforementioned methods by comparing sensor prediction performance and the computational time spent in the design phase. We illustrate the proposed methodology and its benefits on an example from chemical engineering domain.


\section{Notation}
The meaning and notation of the symbols used in the contribution is following: $\bm 1$ denotes a vector of ones, $b_p$ denotes an off-set (constant part) of the inferential sensor, $b_w$ denotes an off-set (constant part) of the separation hyperplane, $e_w$ denotes a vector of slack variables of SVM, $\gamma$ denotes a weighting parameter of SVM, $i$ denotes a measurement index, $I$ denotes an index set for the number of measured data points available for training with cardinality $n$, $I_\text{cl}$ denotes an index set for the number of the classes (models) with cardinality $n_{\text{cl}}$, $I_\text{sp}$ denotes an index set for the number of the separation hyperplanes with cardinality $n_{\text{sp}}$, $p$ denotes a vector of the sensor parameters, $w$ denotes a normal vector to the separation hyperplane, $x$ denotes a vector of easy-to-measure variables, $y$ denotes true value of the desired (output, hard-to-measure) variable, $\hat{y}$ denotes the estimated value of the desired variable, and $Z$ denotes a labeling matrix.

\section{Problem Statement}
Our goal is to identify a multi-model inferential sensor (MIS) in the following piece-wise linear form~\cite{su_2011}:
\begin{equation}
	\hat{y}=
	\begin{cases} 
        p_1^\intercal x + b_{p,1}, & \text{if } x \in \mathcal R_1,\\
        p_2^\intercal x + b_{p,2}, & \text{if } x \in \mathcal R_2,\\
        \ \ldots\\
        p_{n_{\text{cl}}}^\intercal x + b_{p,n_{\text{cl}}}, & \text{if } x \in \mathcal R_{n_{\text{cl}}},
	\end{cases}
\end{equation}
where $\hat{y}$ stands for the inferred value of the desired variable, $x\in\R^{n_p}$ is a vector of the input data, $p_r\in\R^{n_p}$ represents a vector of the sensor parameters in region $\mathcal R_r$, $b_{p,r}$ is a constant sensor off-set, and $n_{\text{cl}}$ is the number of classes/models. We will assume $n_{\text{cl}}$ to be fixed throughout this paper. We note that the number of models is a designer's choice and that this methodology can be applied for the cases with any $n_{\text{cl}}$. Regions of individual model validity denoted as $\mathcal R_r$ represent convex polyhedra such that $\mathcal R_r\bigcap\mathcal R_s=\emptyset, \forall r, s\in I_\text{cl}:=\{1, 2, \ldots, n_{\text{cl}}\}, r\neq s$.

The design of an inferential sensor is possible due to the existence of $n$ measurements of the output, $y\in\R^n$. Next, we describe the standard approach of the MIS design (MIS-std). It consists of three principal steps: data labeling, data classification, individual sensor training.

\subsection{Data labeling of the training dataset}
There are several methods~\cite{maimon_2010} capable of providing accurate data labeling. Here, we use $k$-means clustering~\cite{forgy_1965} within MIS-std. This method separates the $n$ data points to the desired number of clusters while the average distance is minimized of the member point to the center of the cluster.

\subsection{Data classification for switching-logic design}
To design a switching logic between the models of the inferential sensor, i.e., to determine $\mathcal R_r, \forall r\in I_\text{cl}$, data classification is performed. We use here the support vector machines (SVM) approach~\cite{vapnik_1992} with linear separators.

\begin{figure}
	\centering
	\begin{subfigure}[t]{0.45\textwidth}
		\centering
		\includegraphics[width=\textwidth]{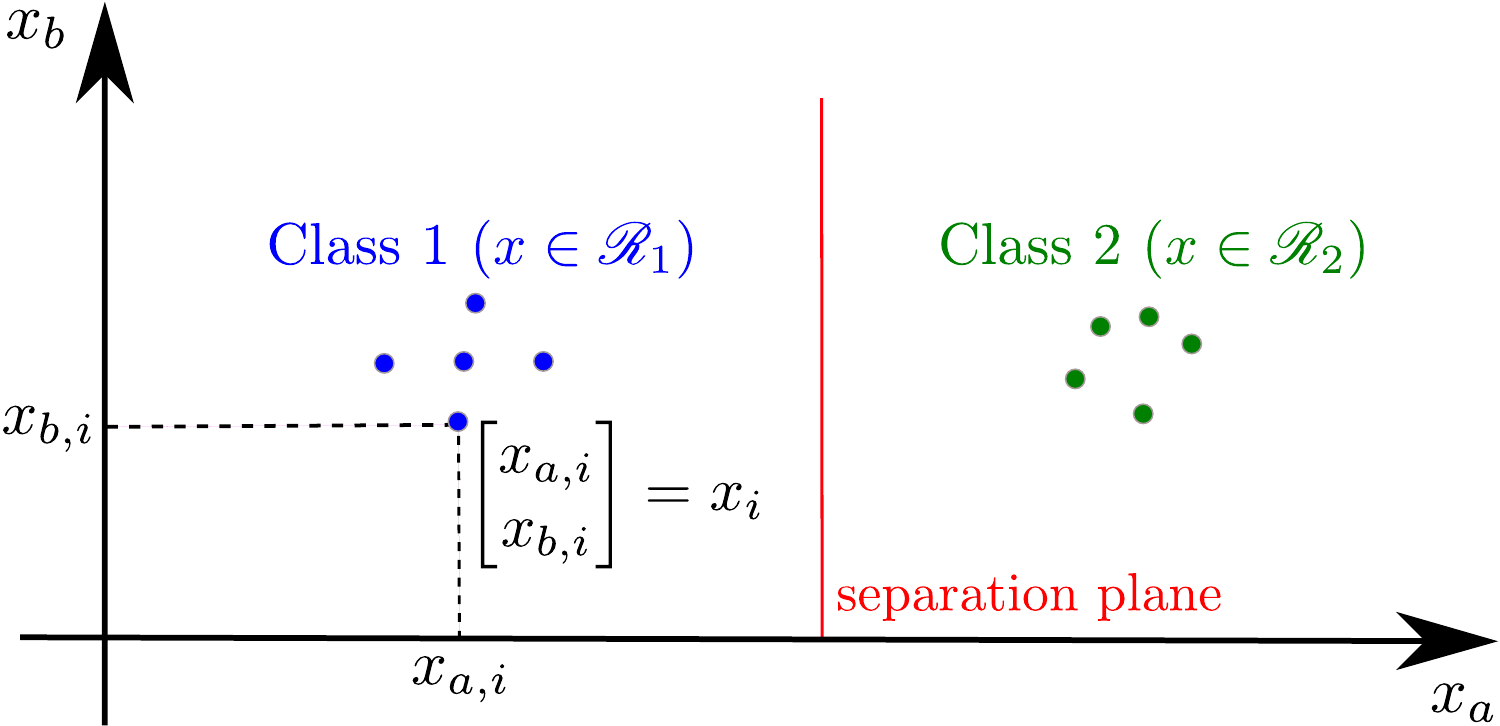}
		\caption{Binary classification.}
		\label{fig:bin_clf}
	\end{subfigure}
	\begin{subfigure}[t]{0.45\textwidth}
		\centering
		\includegraphics[width=\textwidth]{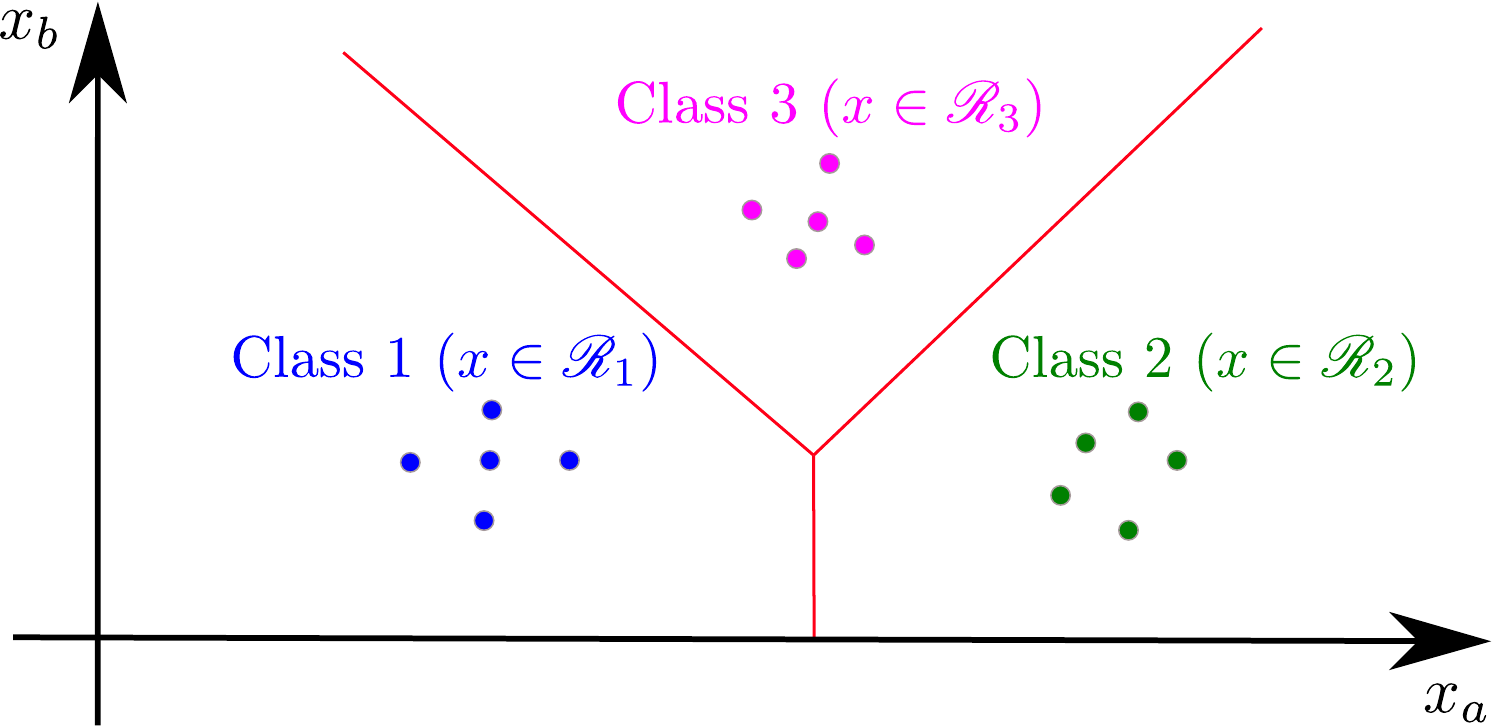}
		\caption{Multi-class classification (three classes).}
		\label{fig:multi_clf}
	\end{subfigure}
	\caption{Types of the data classification problem.}
\end{figure}

According to the number of the desired classes, it is possible to distinguish a binary (see in Fig.~\ref{fig:bin_clf}) or multi-class (see in Fig.~\ref{fig:multi_clf}) classification problem. The multi-class SVM-based classification relies on the use of one-vs-rest or one-vs-one approach~\cite{bishop_2006}, which transforms the multi-class classification into several binary classifications. The one-vs-one method yields more precise classification, yet involves a higher computational burden. It is the method used in this paper, where $n_{\text{sp}} = n_{\text{cl}}(n_{\text{cl}}-1)/2$ separation hyperplanes (binary classificators) are found.

A binary classification~\cite{kecman_2005} can be performed as:
\begin{align}
 &\min_{w, b_w, e\geq0} \ \norm{w}_2 + \gamma\left(\bm 1^\intercal e\right)\\
 &\text{s.t. } z_{i,1}\left(w^\intercal x_i + b_w + e_i - 1\right)\geq 0, \quad \forall i\in I,\\
 &\quad -z_{i,2}\left(w^\intercal x_i + b_w - e_i + 1 \right)\geq 0, \quad \forall i\in I,
\end{align}
where $w$ is the normal vector to the separation hyperplane, $b_w$ is a constant off-set of the separation hyperplane, $\gamma$ is a weighting parameter, and $e$ is a vector of slack variables. The binary parameters $z_{i, j}$ are elements of a labeling matrix $Z\in\R^{n\times n_\text{cl}}$ (clearly, $n_\text{cl}=1$ for binary classification) that results from the data labeling procedure as:
\begin{equation}
 z_{i, j} = 
  \begin{cases}
    1, & \text{if } x_i \in \mathcal R_j,\\
    0, & \text{if } x_i \notin \mathcal R_j.
  \end{cases}
\end{equation}
We define the index set $I_\text{sp}:=\{1, 2, \dots, n_\text{sp}\}$. To ensure classification uniqueness of each data point, matrix $Z$ satisfies
\begin{equation}\label{eq:z_unique}
	\textstyle\sum_{j=1}^{n_\text{cl}} z_{i,j} = 1, \quad \forall i \in I.
\end{equation}

The result of binary classification establishes the switching logic with the regions defined as $\mathcal R_1 :=\{x\in\mathbb R^{n_p}|w^\intercal x_i + b_w\geq0\}$ and $\mathcal R_2 :=\{x\in\mathbb R^{n_p}|w^\intercal x_i + b_w<0\}$.

The one-vs-one multi-class classification can be solved by:
\begin{subequations}\label{eq:svm_1vs1}
 \begin{align}\label{eq:svm_1vs1_obj}
   &\min_{\{w_k, b_{w,k}, e_k\geq0\}_{k\in I_\text{sp}}} \ \textstyle\sum_{k=1}^{n_{\text{sp}}}\norm{w_k}_2 + \gamma\left(\bm 1^\intercal e_{k}\right)\\
   &\text{s.t. } \forall i\in I, \forall k\in I_{\text{sp}}, \text{for }(r,s)= \mathfrak C(n_\text{cl}, k):\notag\\
   & \quad\qquad z_{i,r}\left(w_k^\intercal x_i + b_{w,k} + e_{k,i} - 1\right)\geq 0,\\
   & \quad\quad -z_{i,s}\left(w_k^\intercal x_i + b_{w,k} - e_{k,i} + 1 \right)\geq 0,
 \end{align}
\end{subequations}
where $\mathfrak C(n_\text{cl}, k)$ represents $k$\textsuperscript{th} (non-repetitive) combination of two positive integers $(r, s)$ such that $r<s$ and $r, s\leq n_\text{cl}$. As in the case of binary classification, the separation hyperplanes parameterized by $\{w_k, b_{w,k}\}_{k\in I_\text{sp}}$ define the switching logic.

\subsection{Individual sensor training}
Once switching logic is designed, one can calculate the parameters of the linear models within the structure of MIS. While there are many suitable regression methods~\cite{mojto_2021}, we use the standard least-squares regression here for simplicity.

There two main drawbacks of the MIS-std approach, as presented in this section. Firstly, discontinuity of the inferential sensor can appear at the switching boundaries of the individual models. This can significantly reduce the performance of the controlled system if such a sensor is used in the process monitoring or control. The second issue is that the data labeling is unaware of the prediction accuracy of the resulting sensor and thus it is practically impossible to achieve distribution of the model-validity regions.

\section{Proposed Design Methods for MIS}
\subsection{Design of MIS with Continuous Switching}
We present a novel approach to the MIS design (MIS-con) that effectively combines the SVM-based data classification with the individual sensor training and that ensures continuity of MIS at the switching boundaries of the individual models. The underling optimization problem reads as:
\begin{subequations}\label{eq:mis_con}
 \begin{align}\label{eq:svm_con_obj}
    &\min_{\substack{\{w_k, b_{w,k}, e_k\geq0\}_{k\in I_{\text{sp}}}\\\{p_j, b_{p,j}\}_{j\in I_{\text{cl}}}}}\sum_{j=1}^{n_{\text{cl}}}\sum_{i=1}^{n}z_{i,j}\left(y_i - p_j^\intercal x_i - b_{p,j}\right)^2\\
 	&\text{s.t. } \forall i\in I, \forall k\in I_{\text{sp}}, \text{for }(r,s) = \mathfrak C(n_\text{cl}, k):\notag\label{eq:mis_con_svm_const}\\
 	& \quad\qquad z_{i,r}\left(w_k^\intercal x_i + b_{w,k} + e_{k,i} - 1\right)\geq 0,\\
    & \quad\quad -z_{i,s}\left(w_k^\intercal x_i + b_{w,k} - e_{k,i} + 1 \right)\geq 0,\\
 	\label{eq:mis_con_ctd_const}
 	& \quad\qquad p_r - p_s - w_k = 0.
 \end{align}
\end{subequations}
The constraint~\eqref{eq:mis_con_ctd_const} ensures continuity at the switch between any two models of the designed MIS. This is achieved by establishing the intersection of model surfaces of any two models to coincide with the switching hyperplane determined in SVM-like fashion. One can also consider further regularization of this problem by introducing elements from original SVM objective~\eqref{eq:svm_1vs1_obj} into~\eqref{eq:svm_con_obj}.

\subsection{Design of MIS with Optimized Data Labeling}
We propose the following approach (MIS-con-lab) based on the MIS-con approach that can significantly reduce the inaccuracies caused by the a priori labeling of the training dataset. This approach searches directly for the optimal data labeling by adding $Z$ among the optimized variables in. The underlying optimization problems reads as:
\begin{subequations}\label{eq:mis_con_lab}
 \begin{align}\label{eq:mis_con_lab_obj}
 	&\min_{\substack{\{w_k, b_{w,k}, e_k\geq0\}_{k\in I_{\text{sp}}}\\Z, \{p_j, b_{p,j}\}_{j\in I_{\text{cl}}}}}\sum_{j=1}^{n_{\text{cl}}}\sum_{i=1}^{n}z_{i,j}\left|y_i - p_j^\intercal x_i - b_{p,j}\right|\\
 	&\text{s.t. } \forall i\in I, \forall k\in I_{\text{sp}}, \text{for }(r,s) = \mathfrak C(n_\text{cl}, k):\notag\\
 	& \qquad \text{Eqs.~\eqref{eq:z_unique}, \eqref{eq:mis_con_svm_const}--\eqref{eq:mis_con_ctd_const}}.
 \end{align}
\end{subequations}
We adopt the sum of absolute errors criterion in the objective function~\eqref{eq:mis_con_lab_obj} in order to reduce the complexity of the optimization problem as this can be transformed to mixed-integer linear program (MILP). If the sum of squared errors is used in the objective function, the optimization problem turns into mixed-integer nonlinear program (MINLP) that might be challenging especially when number of available training points $n$ is high.

Note that the problem~\eqref{eq:mis_con_lab} serves primarily to decide about data labels, i.e., distribution of the training data and, subsequently, of model validity regions. After fixing the values of $Z$, the final training can be performed via solving~\eqref{eq:mis_con} so that the final models are trained using the same (least-squares) criterion. This approach no longer requires the a priori labeling of the training dataset and can provide optimal MIS. The prize to pay is the increased computational burden. 

The problem~\eqref{eq:mis_con_lab} can be transformed to quadratically constrained quadratic program (MIQCQP) using the epigraph form~\cite{milano_2012} of the absolute value in~\eqref{eq:mis_con_lab_obj}. Arising bilinear constraints can be further transformed to linear by using big-M method~\cite{griva_2008}. As the variables in $Z$ are already binary, the big-M method does not require any new integer variables. The resulting form of MIS-con-lab is thus an MILP.

\section{Case Study}
The performance of the MIS design methods is tested for the estimation of pressure compensated temperature $PCT$. This is a phenomenological variable used very often in the petrochemical industry. It is derived by combining the Antoine and Clausius-Clapeyron equations as~\cite{king_2011}:
\begin{equation}
	\frac{1}{PCT} = \frac{R}{H_v}\log\left(\frac{P}{P_{\text{ref}}}\right) + \frac{1}{T},
\end{equation}
where $H_v$ is the heat of vaporization, $R$ is the universal gas constant, $P_{\text{ref}}$ is the reference pressure, $P$ is an absolute pressure and $T$ is an absolute temperature.

We assume no knowledge about the structure of the $PCT$ model. The only information available is given by the 90 measurements of $PCT$ at different temperatures and pressures. The ground-truth values of parameters of the $PCT$ model are $R = 8.314\,\text{J/mol/K}$, $H_v = 55,940.550\,\text{J/mol}$, $P_{\text{ref}}  = 145,325\,\text{Pa}$. The data used in the simulations is generated within the following intervals:
\begin{subequations}
	\begin{align}
	    2,000\,\text{Pa} \leq P & \leq 20,000\,\text{Pa},\\
	    523.15\,\text{K} \leq T & \leq 573.15\,\text{K}.
	\end{align}
\end{subequations}
We further assume that the training samples of $PCT$ are corrupted with some noise. This noise is generated as a random variable from the standard normal distribution.

In order to remove the discrepancies in the variables magnitudes ($P$, $T$, $PCT$), we performed a normalization. Therefore, the normalized variables ($P_{\text{norm}}$, $T_{\text{norm}}$, $PTC_{\text{norm}}$) used in the further experiments are within the interval $[0,1]$.

We will study two situations regarding the distribution of the a priori available training data: one with three linearly separable clusters of data points and the other with no apparent clusters (data points randomly scattered in the training area). In both cases, we seek a three-model sensor.

We compare the performance of single-model inferential sensor (SIS), MIS-std, MIS-con, and MIS-con-lab. Except for MIS-con-lab that is identified using BARON~\cite{sahinidis_2017}, all the designs are found using Gurobi~\cite{gurobi_2021}. 
All the solvers are limited to at most one hour of run time. Alternatively to solving MIS-con-lab using BARON (denoted as MIS-con-lab (bar)), we use BARON with its heuristic termination option and denote the approach as MIS-con-lab (bar,ht). When the MINLP problem~\eqref{eq:mis_con_lab} is transformed into MILP, the resulting problem is solved with Gurobi (denoted as MIS-con-lab (gur)).

\begin{figure}
	\centering
	\includegraphics[width=0.48\textwidth]{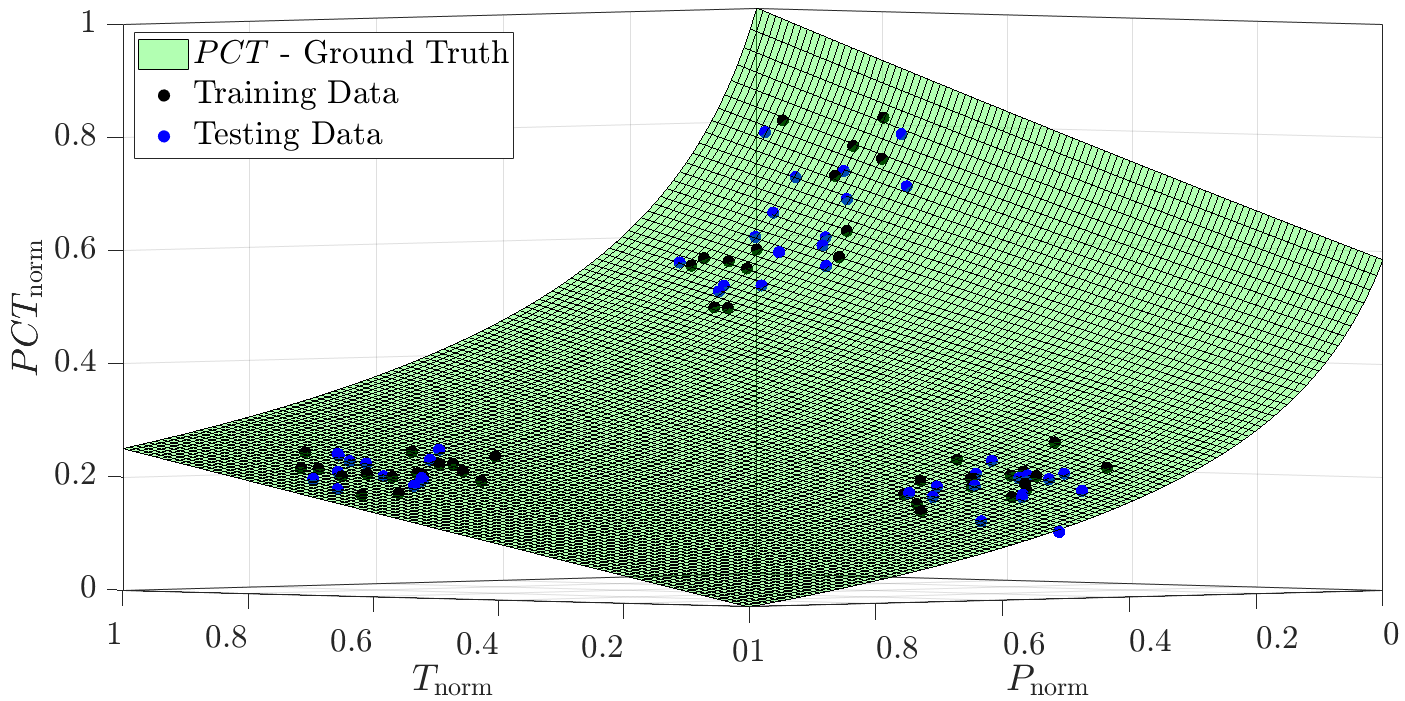}
	\caption{The $PCT$ model with clustered dataset.}
	\label{fig:pct_ls}
\end{figure}

\begin{figure}
	\centering
	\includegraphics[width=0.48\textwidth]{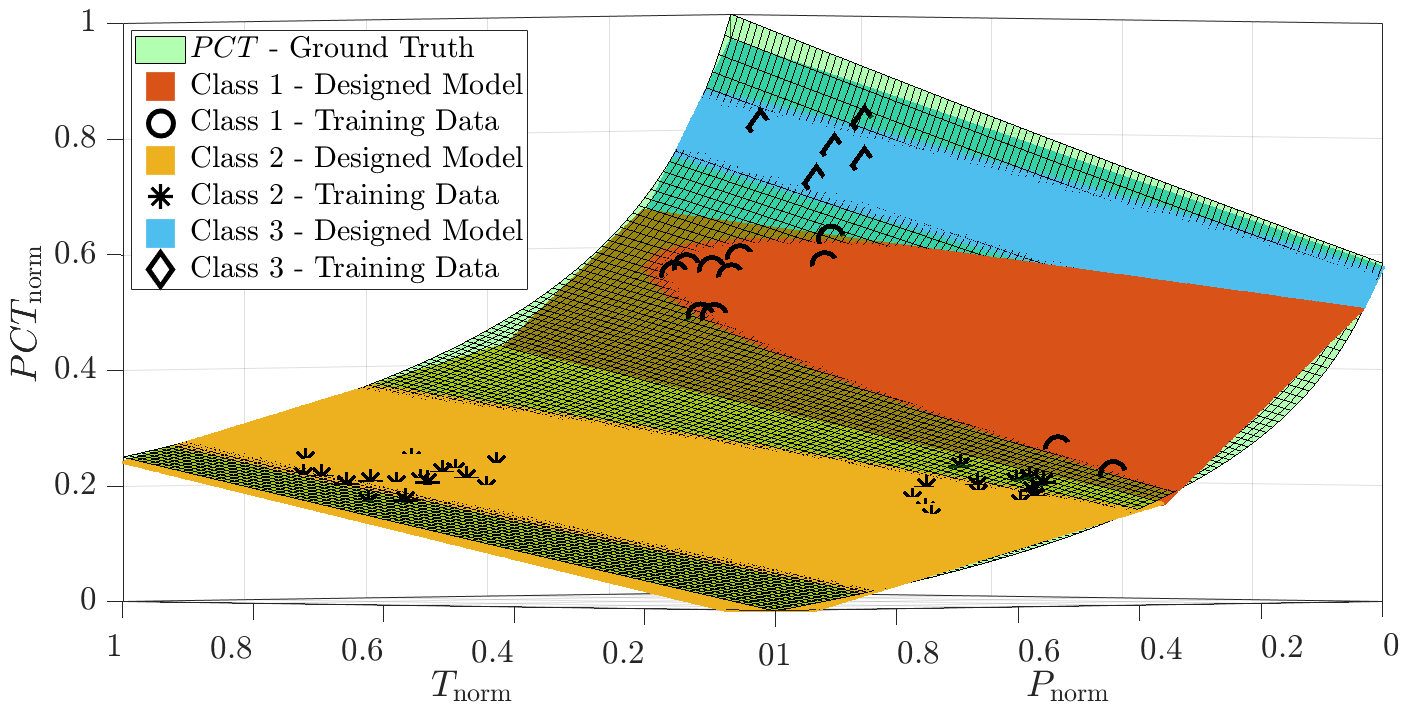}
	\caption{The approximated $PCT$ model on the clustered training dataset by using MIS-con-lab (gur).}
	\label{fig:pct_model_linsep}
\end{figure}

\subsection{Clustered Dataset}
In this simulated experiment, we design the MIS using clustered data points. This situation mimics the sensor training when industrial data is well treated and distinct operating points are evident. As we can see in Fig.~\ref{fig:pct_ls}, the data points are concentrated into three clusters. Each cluster contains 30 data points. Subsequently, we randomly assign 50~\% of the data to the training set (black points) and the remaining data to the testing set (blue points). 

The approximation of the original $PCT$ model by MIS-con-lab (gur) is shown in Fig.~\ref{fig:pct_model_linsep}. The training dataset was optimally labeled into the different classes and individual models were trained (class 1 - circles, red model surface; class 2 - stars, yellow model surface; class 3 - diamonds, blue model surface). Models with non-optimized sampling fall for establishing the model validity regions based on the (a priori) visible clusters. The MIS-con-lab (gur) seems to be quite accurate on the training dataset.

\begin{figure}
	\centering
	\includegraphics[width=0.48\textwidth]{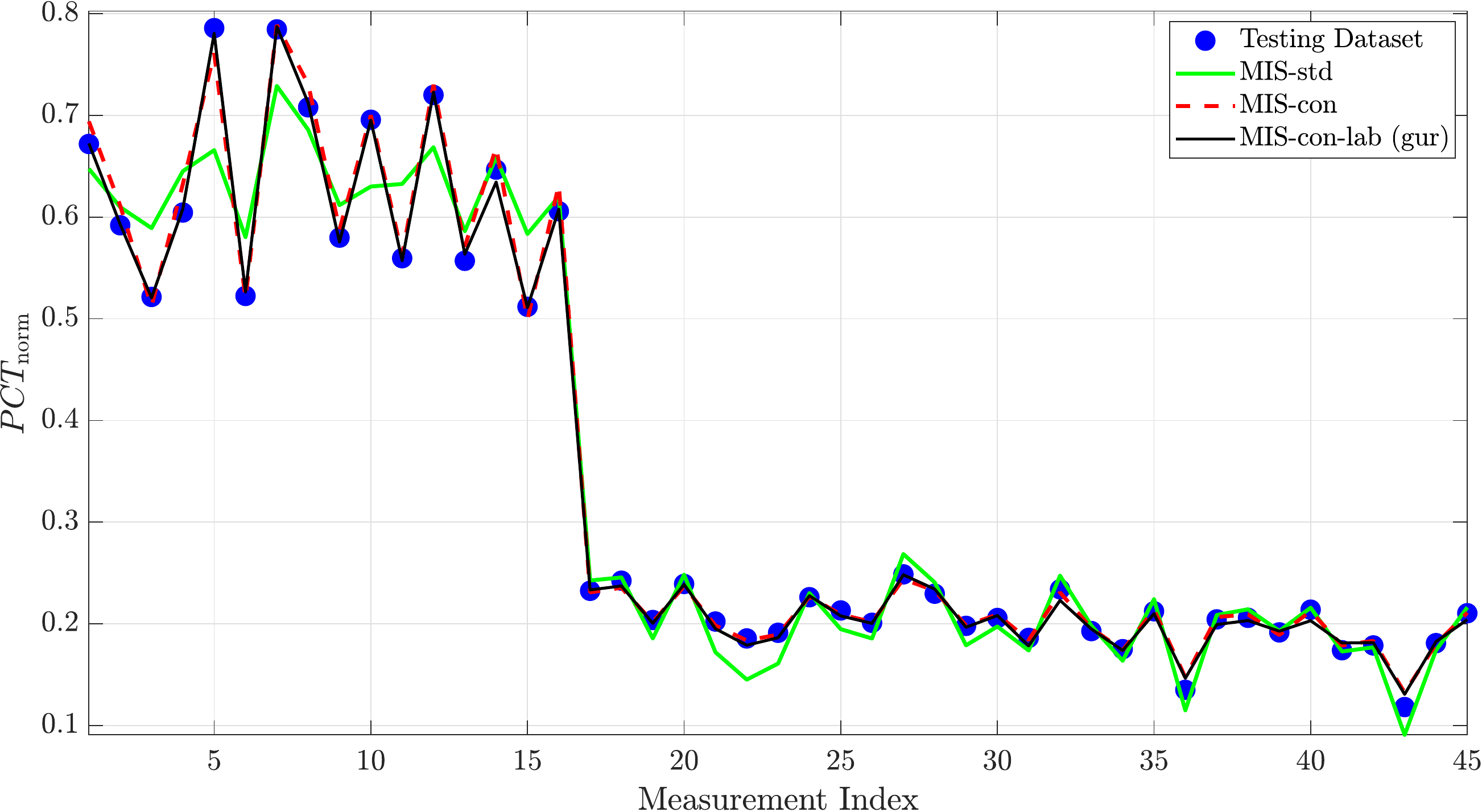}
	\caption{The prediction performance of the designed inferential sensors on the clustered testing dataset.}
	\label{fig:pct_ls_ts}
\end{figure}

We further compare the accuracy of the designed MISs on the testing dataset. This comparison is shown in Fig.~\ref{fig:pct_ls_ts}. As we can see, the accuracy of the standard approach MIS-std (solid green line) is lower compared to the MIS-con
(dashed red line) approach or the MIS-con-lab (gur) approach (solid black line). The largest discrepancy between the approaches is visible within the interval with highest values of $PCT_{\text{norm}}$ (measurements 1--16) located in the part of most pronounced nonlinearity of the $PCT$ model (see in Fig.~\ref{fig:pct_ls}). The reason for this discrepancy is that the proposed approaches try to effectively fit the nonlinear section of $PCT$ by two models (see in Fig.~\ref{fig:pct_model_linsep}) while the MIS-std approach uses only one model in this area.

\begin{table}
	\caption{The accuracy (RMSE) and computational time ($t_{\text{comp}}$) of the studied approaches on the clustered dataset.}
	\centering
	\begin{tabular}{l|c|c|c}
		Approach & \begin{minipage}{1.7cm}\centering RMSE $\times 10^2$\\Training Data\end{minipage} 
		& \begin{minipage}{1.7cm}\centering RMSE $\times 10^2$\\Testing Data\end{minipage} 
		& \begin{minipage}{1cm}\centering $t_{\text{comp}}$ [s]\end{minipage}\\\hline
		                     &        &        &     \\[-0.2cm]
		SIS                  & 4.1 & 3.5 & 0.1   \\
		MIS-std              & 0.9 & 1.0 & 4.3   \\
		MIS-con              & 0.9 & 1.1 & 4.1   \\
		MIS-con-lab (bar)    & 0.5 & 1.6 & 3,600.0  \\
		MIS-con-lab (bar,ht) & 0.5 & 1.6 & 325.9 \\
		MIS-con-lab (gur)    & 0.4 & 0.5 & 2,284.0
	\end{tabular}
	\label{tab:summary_case_1}
\end{table}

The resulting values of the accuracy measured by the root-mean squared error (RMSE) and computational time ($t_{\text{comp}}$) for each studied approach are shown in Tab.~\ref{tab:summary_case_1}. The values of the accuracy of designed MISs confirm the best performance of the MIS-con-lab (gur) approach that we could indicate in Fig.~\ref{fig:pct_ls_ts} as well. The MIS-con-lab (bar) approach is constrained by the time limit (3,600 seconds), and therefore this approach achieved worse accuracy on the testing dataset than other MIS approaches. From the computational time perspective (see in Tab.~\ref{tab:summary_case_1}), the proposed MIS-con approach is comparable with the MIS-std approach. It seems that the heuristic termination of the BARON can reduce this time burden while it provides the MIS with the same accuracy as the MIS-con-lab (bar) approach.

\subsection{Uniformly Distributed Dataset}
\begin{figure}
	\centering
	\includegraphics[width=0.48\textwidth]{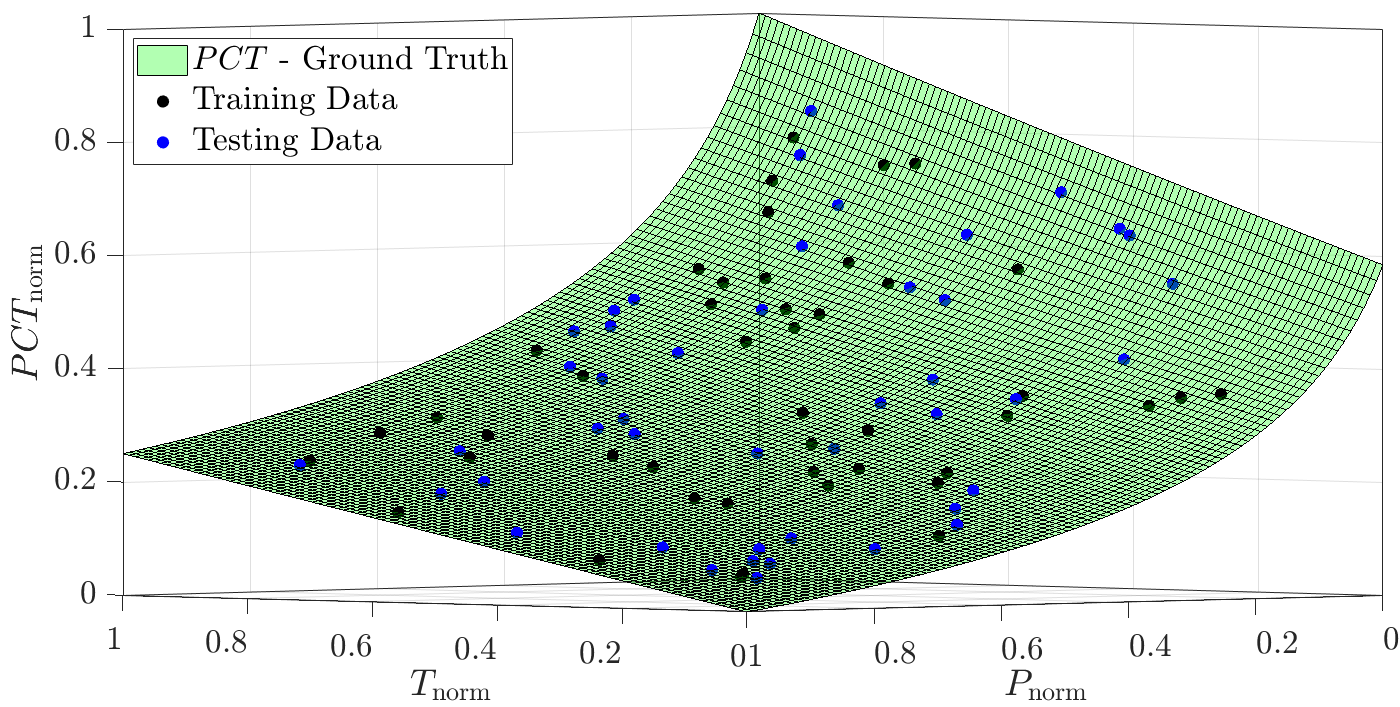}
	\caption{The $PCT$ model with uniformly distributed dataset.}
	\label{fig:pct_nls}
\end{figure}

In this simulated experiment, the data points are not concentrated into clusters but uniformly distributed within the operation interval (see Fig.~\ref{fig:pct_nls}). Unlike the previous experiment, this situation resembles datasets seen more often in industrial practice. We randomly assign 50~\% of the data to the training set (black points) and the remaining data to the testing set (blue points). Due to the random character of this scenario, we perform 100 simulation runs with different uniformly distributed datasets. Subsequently, we statistically evaluate the achieved results.

\begin{figure}
	\centering
	\includegraphics[width=0.48\textwidth]{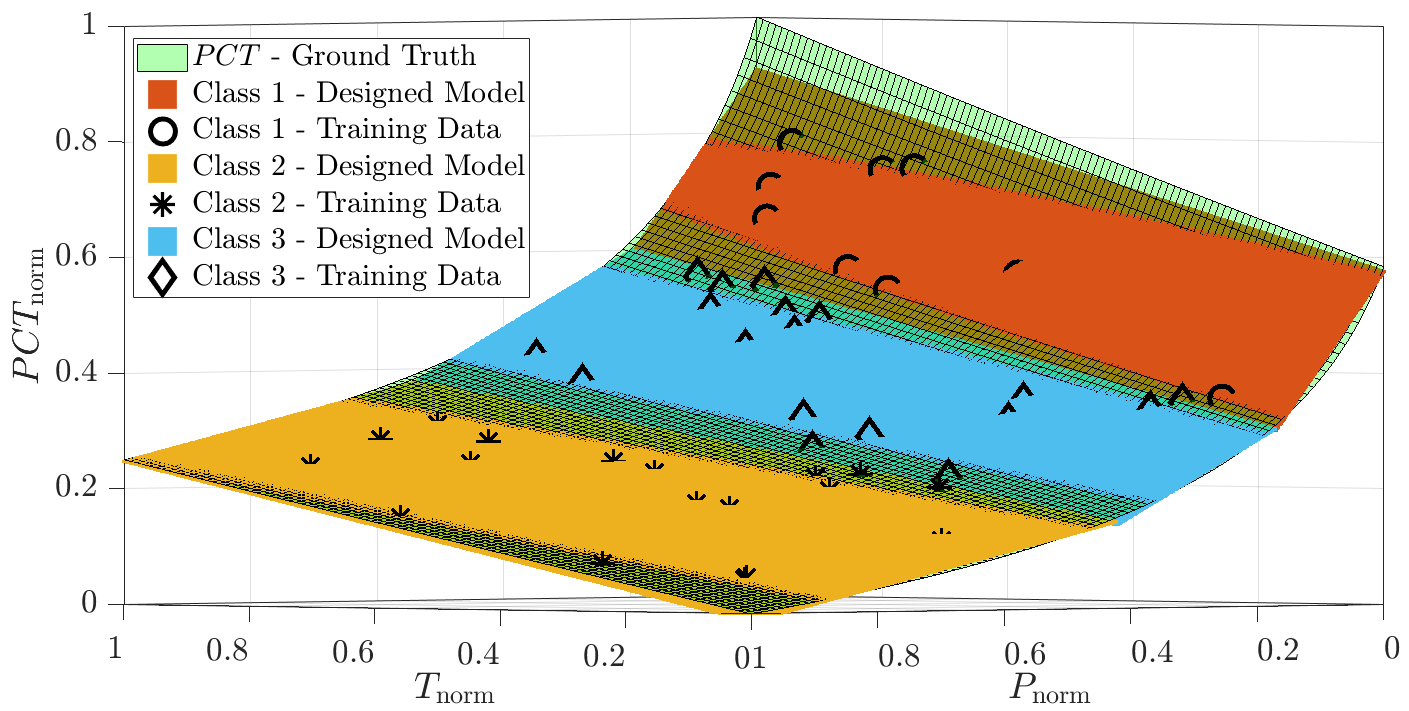}
	\caption{The approximated $PCT$ model on the uniformly distributed testing dataset by using MIS-con-lab (gur).}
	\label{fig:pct_model_nlinsep}
\end{figure}

Before analyzing the results from 100 simulation runs, we look at the results from one representative run. The approximation of the ground-truth model by MIS-con-lab (gur) approach is shown in Fig.~\ref{fig:pct_model_nlinsep}. We illustrate this approach because of its better performance (higher accuracy) on the training dataset compared to other approaches. The results in Fig.~\ref{fig:pct_model_nlinsep} show designed models, where same color and symbol code is used as in the case of clustered dataset. Due to the minimal discrepancy between the designed MIS and the ground-truth model in Fig.~\ref{fig:pct_model_nlinsep}, we can conclude that the proposed MIS-con-lab (gur) approach can accurately approximate the $PCT$ model within the concerned training dataset.

\begin{figure}
	\centering
	\includegraphics[width=0.48\textwidth]{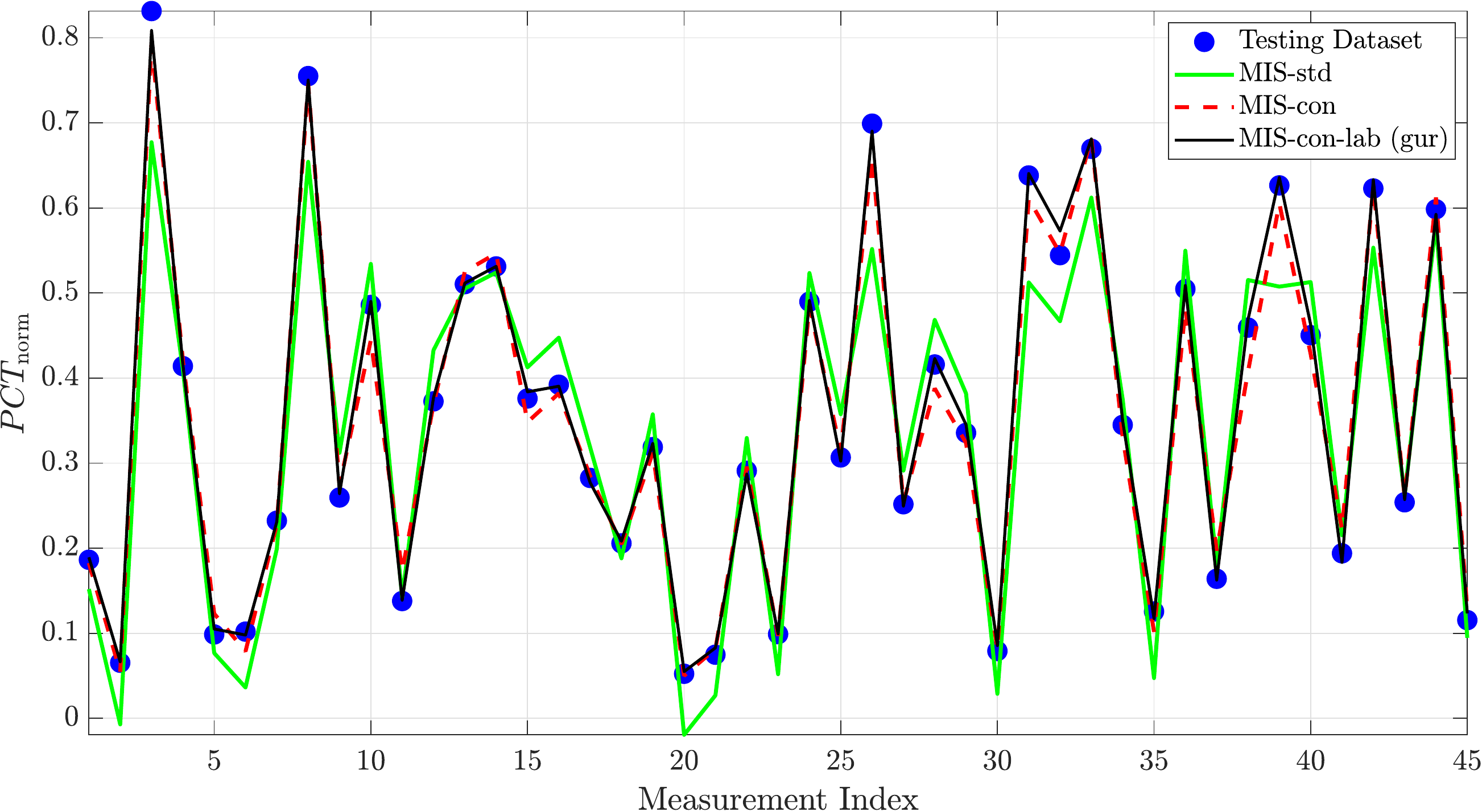}
	\caption{The prediction performance of the designed inferential sensors on the uniformly distributed testing dataset.}
	\label{fig:pct_nls_ts}
\end{figure}

According to Fig.~\ref{fig:pct_nls_ts}, which shows an accuracy comparison of the chosen designed MISs, we can see that the MIS-std approach is less accurate on the testing dataset than the MIS-con approach and the MIS-con-lab (gur) approach. The accuracy of the MIS-con and MIS-con-lab (gur) approaches seems to be similar. However, there are testing measurements (3, 8, 11, 28, 31, and 39) in Fig.~\ref{fig:pct_nls_ts} which confirm the highest accuracy of the MIS-con-lab (gur) approach. Due to the location of these measurements, we conclude again that the MIS-con-lab (gur) approach is capable of better approximating the nonlinear section of the PCT model compared to other approaches.

\begin{figure}
	\begin{minipage}{\linewidth}
		\centering
		\includegraphics[width=\linewidth]{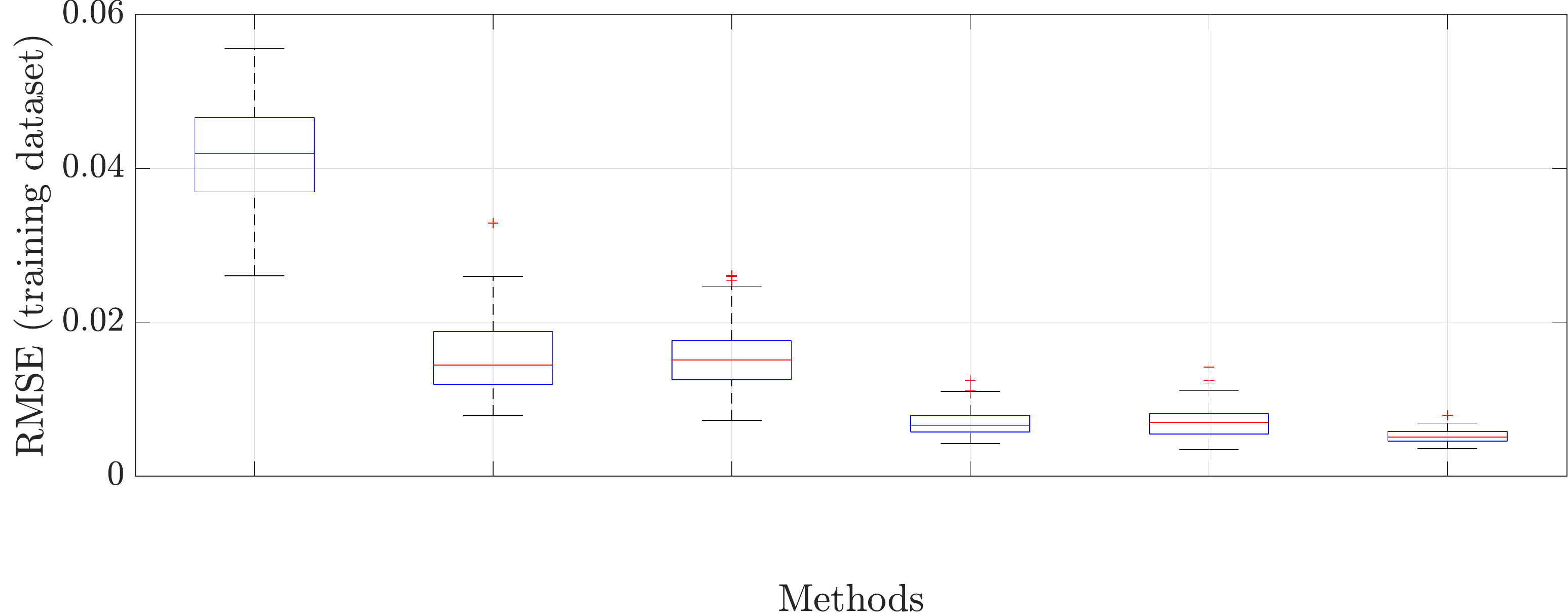}\vspace{-.6cm}
	\end{minipage}
	\begin{minipage}{\linewidth}
		\centering
		\includegraphics[width=\linewidth]{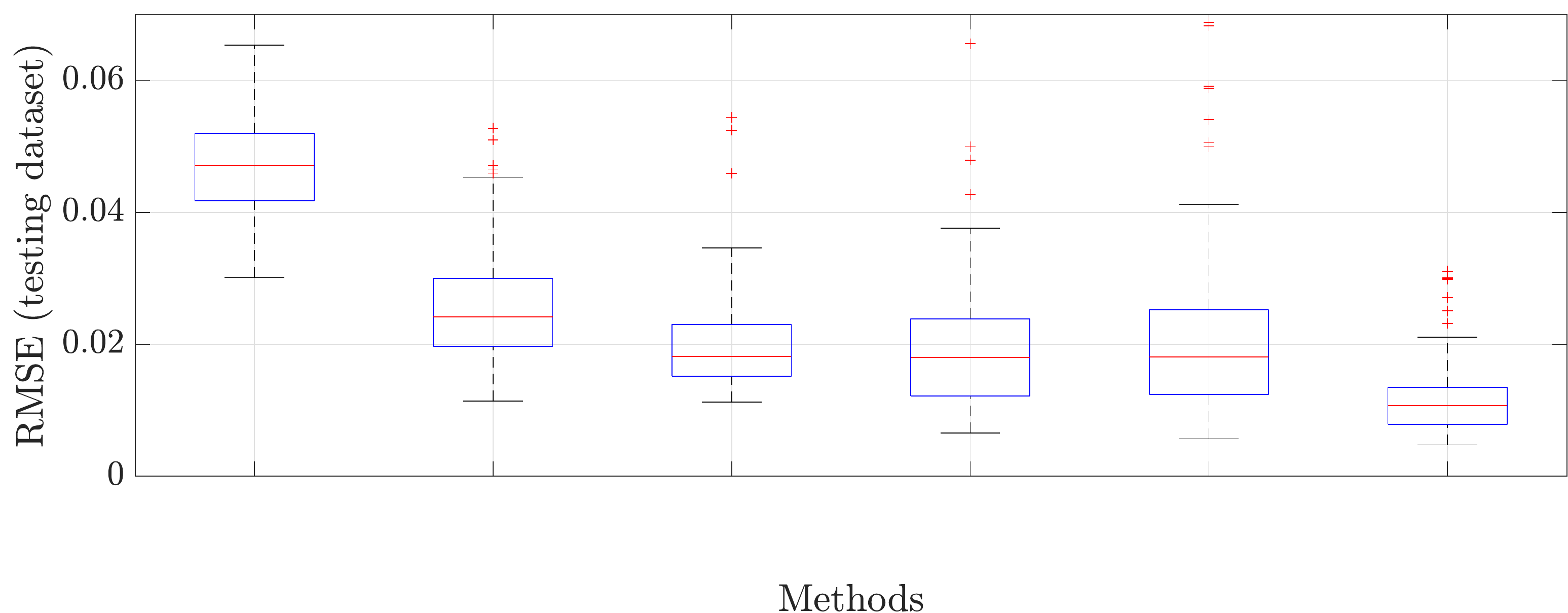}\vspace{-.6cm}
	\end{minipage}
	\begin{minipage}{\linewidth}
		\includegraphics[width=\linewidth]{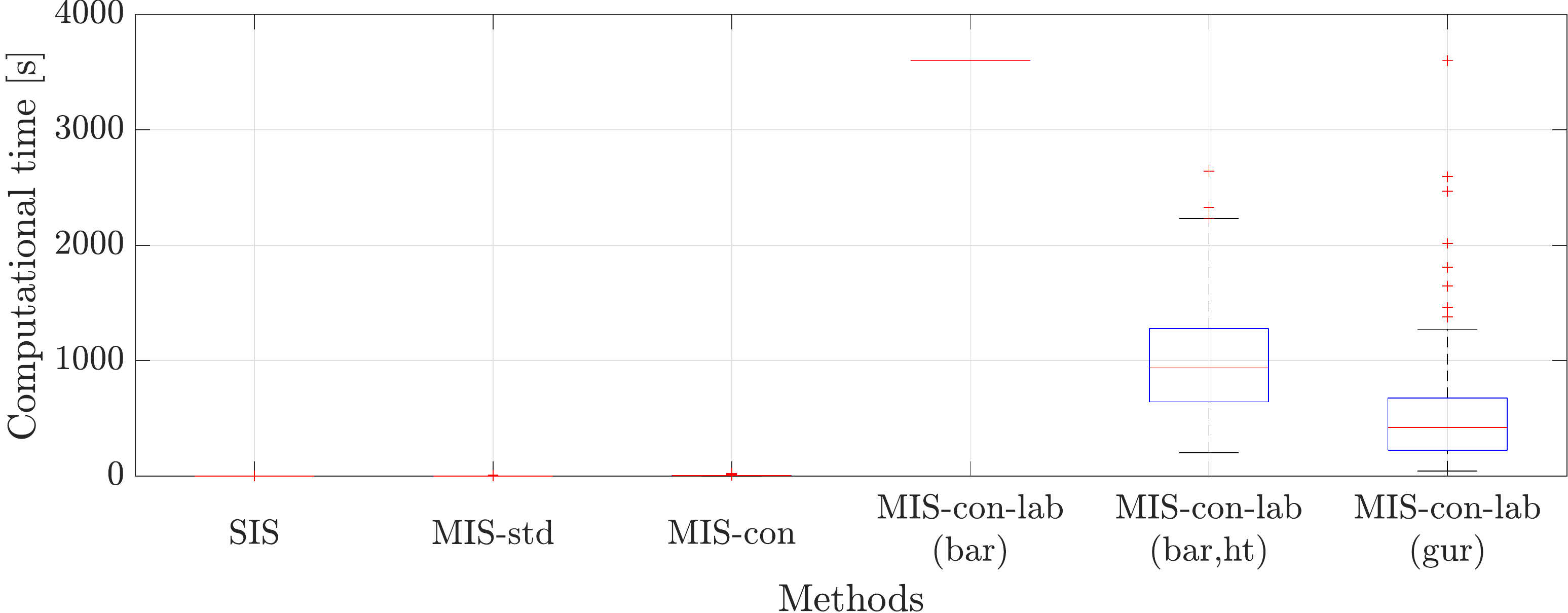}
	\end{minipage}
	\caption{The accuracy (RMSE) of the designed inferential sensors on the training (top) and testing datasets (middle) and computation time of each method (bottom) from 100 simulation runs with different uniformly distributed datasets.}
	\label{fig:pct_nls_boxplots}
\end{figure}

We further statistically analyze the results evaluated on 100 runs with different uniformly distributed datasets. The results are shown in Fig.~\ref{fig:pct_nls_boxplots}, where the blue box represents 25\textsuperscript{th} and 75\textsuperscript{th} (indicating variance), the red line inside the blue box is a median, and red crosses outside of the blue box are outliers. We can conclude (see in Fig.~\ref{fig:pct_nls_boxplots}, top and middle plots) that the single-model inferential sensor (SIS) is far less accurate than the designed multi-model inferential sensors (MIS). According to the accuracy of designed MISs, we can directly conclude that the MIS-con-lab (gur) approach is the most accurate one on both training and testing datasets. The comparison of the variances (see blue boxes in Fig.~\ref{fig:pct_nls_boxplots}, top and middle plots) confirms the best performance of the MIS-con-lab (gur) approach. The large variance and the amount of the present outliers within the results from the testing dataset confirm that the MIS-con-lab approach solved with BARON exhbits problems in identifying global optima (thre is a tendency of over-fitting), which significantly affects the prediction accuracy evaluated on the testing dataset.

According to the results from the computational time (see in Fig.~\ref{fig:pct_nls_boxplots}, bottom plot), we can conclude that the MIS-std and MIS-con approaches require lower computational time than the rest of the methods. Furthermore, we can see that the computational time of MIS-con-lab (bar,ht) is significantly reduced compared to MIS-con-lab (bar). Nevertheless, the accuracy of designed sensors is comparable. We could make similar conclusions as in the previous scenario. Moreover, the results from the bottom plot in Fig.~\ref{fig:pct_nls_boxplots} indicate improvement of the computational time of the MIS-con-lab (gur) compared to the previous scenario (see in Tab.~\ref{tab:summary_case_1}). It seems that this approach is more suited for well distributed measurements over the operational space than the data concentrated in the clusters. This is caused by the fact that such distributed data points provide only a small space for the variation of the results (less local over-fitting) of the MIS-lab-con (gur) approach.

\section{Conclusions}
In this paper, we presented novel approaches for designing multi-model linear inferential sensors that ensure continuity at switching between models and that allow for optimized data labeling of the training dataset. While the optimal data labeling involves computationally intensive problem (MINLP), we provided a transformation that yields an MILP form. The performance of the studied approaches was evaluated according to two scenarios. The first scenario involved the clustered data, while the second scenario involved uniformly distributed data. The results from both scenarios indicate the best accuracy of the MIS-con-lab (gur) approach on the training and testing datasets. The price for the achieved accuracy is a higher computational burden that resulted in significantly longer computational time than the MIS-std approach, yet the achieved 50\,\% accuracy improvement might be worthy of consideration. Our future work will involve the introduction of more advanced training methods (LASSO or subset selection) within the MIS design.


\section*{Acknowledgement}
We acknowledge discussions with M. Klaučo and K. Kiš from STUBA and with K. Ľubušký from Slovnaft, a.s.

\bibliographystyle{IEEEtran}
\bibliography{IEEEabrv,references}

\end{document}